\documentclass[runningheads]{llncs}

\usepackage{graphicx}
\usepackage{url}
\begin{document}
\title{Automating Turkish Educational Quiz Generation Using Large Language Models\thanks{The funding for this paper was provided by the TAILOR project and the HumanE-AI-Net projects, both supported by the EU Horizon 2020 research and innovation program under GA No 952215 and No 952026, respectively.}}
%
%
\author{ Kamyar Zeinalipour\inst{1}\orcidID{0009-0006-3014-2511} \and Yusuf Gökberk Keptiğ\inst{1}\orcidID{0009-0002-2793-9061} \and Marco Maggini\inst{1}\orcidID{0000-0002-6428-1265} \and Marco Gori\inst{1}\orcidID{0000-0001-6337-5430}}

\authorrunning{Zeinalipour et al.}
%
\institute{University of Siena, Siena, Italy\\ \email{\{kamyar.zeinalipour2, marco.maggini, marco.gori\}@unisi.it
} \email{y.keptig@student.unisi.it}\ }
\maketitle              
\begin{abstract}
Crafting quizzes from educational content is a pivotal activity that benefits both teachers and students by reinforcing learning and evaluating understanding. In this study, we introduce a novel approach to generate quizzes from Turkish educational texts, marking a pioneering endeavor in educational technology specifically tailored to the Turkish educational context. We present a specialized dataset, named the \textit{Turkish-Quiz-Instruct}, comprising an extensive collection of Turkish educational texts accompanied by multiple-choice and short-answer quizzes. This research leverages the capabilities of Large Language Models (LLMs), including \textit{GPT-4-Turbo}, \textit{GPT-3.5-Turbo}, \textit{Llama-2-7b-chat-hf}, and \textit{Llama-2-13b-chat-hf}, to automatically generate quiz questions and answers from the Turkish educational content. Our work delineates the methodology for employing these LLMs in the context of Turkish educational material, thereby opening new avenues for automated Turkish quiz generation. The study not only demonstrates the efficacy of using such models for generating coherent and relevant quiz content but also sets a precedent for future research in the domain of automated educational content creation for languages other than English. The \textit{Turkish-Quiz-Instruct} dataset is introduced as a valuable resource for researchers and practitioners aiming to explore the boundaries of educational technology and language-specific applications of LLMs in Turkish. By addressing the challenges of quiz generation in a non-English context specifically Turkish, this study contributes significantly to the field of Turkish educational technology, providing insights into the potential of leveraging LLMs for educational purposes across diverse linguistic landscapes.

\keywords{Turkish Quiz Generator  \and Large Language Models \and LLMs \and Automatic Generation \and Question Generation  \and Multiple-Choice Questions \and Short-Answer Questions \and Quize Generation }

\end{abstract}
\section{Introduction}

In recent years, the landscape of education has been dramatically transformed by the integration of technology, enabling innovative methods to enhance learning quality and effectiveness. Among these technological interventions, quizzes have stood out as valuable tools in the educational process. They serve a dual purpose: reinforcing learning through practice and feedback, and evaluating students' understanding of the subject matter. This has sparked a growing interest among educators and researchers in optimizing quiz generation, specifically in how technology can be leveraged to streamline and improve this process. The present study embarks on exploring this avenue within the Turkish educational context—a linguistic and cultural niche previously underexplored in the realm of educational technology.\cite{butler2018multiple,towns2014guide,dengri2021review}. \\
The crafting of quizzes from educational content, particularly in languages other than English, presents a unique set of challenges and opportunities. The Turkish language, with its unique syntax and semantic structures, poses specific complexities that require tailored solutions for effective quiz generation. Recognizing this gap, our study introduces an approach to addressing these challenges, marking a significant contribution to the field of educational technology. By focusing on the Turkish educational context, we not only cater to the immediate needs of Turkish learners and educators but also enrich the broader discourse on language-specific educational technology solutions.\\
Our contribution to this evolving field of study is twofold. Firstly, we present the \textit{Turkish-Quiz-Instruct}\footnote{\url{https://huggingface.co/datasets/Kamyar-zeinalipour/Turkish-Quiz-Instruct}} dataset—a comprehensive collection of Turkish educational texts paired with meticulously crafted multiple-choice and short-answer quiz questions. This dataset emerges as a pioneering resource, providing a robust foundation for the development and testing of quiz generation models tailored to the Turkish language. Secondly, we exploit the capabilities of contemporary Large Language Models (LLMs), specifically \textit{GPT-3.5-Turbo},  \textit{Llama-2-7b-chat-hf}, and  \textit{Llama-2-13b-chat-hf} to automate the generation of Turkish quiz questions and answers from the educational content by fine-tuning on the introduced dataset.\\ 
By evaluating the results of the generated quizzes, this research endeavors to demonstrate the efficacy and reliability of employing LLMs after fine-tuning in the Turkish educational landscape. In doing so, it lays the groundwork for further exploration and development in the field, hinting at the vast potential of tailored educational technology solutions across various languages and cultures. The introduction of the \textit{Turkish-Quiz-Instruct} dataset and the investigation of different models\footnote{\url{https://huggingface.co/Kamyar-zeinalipour/TR_QUIZ_GEN_SIMPLE_LLAMA7B}\\ \url{https://huggingface.co/Kamyar-zeinalipour/TR_QUIZ_GEN_SIMPLE_LLAMA13B}\\
\url{https://huggingface.co/Kamyar-zeinalipour/TR_QUIZ_GEN_MULTI_LLAMA7B}\\ \url{https://huggingface.co/Kamyar-zeinalipour/TR_QUIZ_GEN_MULTI_LLAMA13B} \\
\url{https://github.com/KamyarZeinalipour/Turkish_Quiz_Generator}}
for Turkish quiz generation significantly enriches the toolkit available to researchers and practitioners in the field. This study not only sets a precedent in the Turkish context but also contributes valuable insights and resources for the broader community engaged in leveraging educational technology for enhanced learning experiences.\\
The organization of this paper is as follows: Section~\ref{sec:relatedworks} surveys the pertinent literature. Section~\ref{sec:method} elaborates on our methodology. Section~\ref{sec:experiments} extensively discusses the outcomes of our experiments. Finally, Section~\ref{sec:conclusions} summarizes our conclusions.

\section{Related Work}~\label{sec:relatedworks}

Question generation (QG) is a crucial task in natural language processing that aims to automatically create a question from a given sentence or paragraph, often with the assistance of answer information when available. The challenge in QG lies in identifying the key statement within the context and generating a question based on that statement. This process can be categorized into answer-aware QG, where the question is generated with knowledge of the answer, or answer-agnostic QG, where the question is generated without knowledge of the answer \cite{dugan2022feasibility}.\\
QG and question answering (QA) are interconnected tasks \cite{tang2017question} that require reasoning between questions and answers. As a result, datasets originally designed for QA tasks, such as SciQ, RACE, and FairytaleQA \cite{welbl2017crowdsourcing,lai2017race,xu2022fantastic}, are also utilized in QG research \cite{wu2023towards,jia2021eqg,steuer2022investigating,zhao2022educational}. Additionally, there are specific datasets created for QG, including LearningQ, KHANQ, and EduQG, which cover a variety of subjects and education levels \cite{chen2018learningq,gong2022khanq,hadifar2023eduqg}. LearningQ contains questions crafted by instructors that demand reasoning but lack answers, limiting its application to answer-aware QG. KHANQ provides triples consisting of context, prompt, and question collected from questions about online courses, with the prompt offering background knowledge or presuppositions. EduQG is a high-quality multiple-choice question dataset linked to cognitive complexity, enhancing its realism.\\
Early QG methods relied on rule matching, but advancements led to the adoption of Seq2Seq models with attention, linguistic feature integration, multi-modal models, multi-task learning, reinforcement learning, and the application of language models like \textit{BERT} and \textit{GPT-3} 
\cite{kunichika2004automated,du2017identifying,harrison2018neural,zhou2018neural,naeiji2022question,wang2023multiqg,zhou2019multi,chen2019natural,chan2019recurrent,wang2022towards}. To incorporate answer information, researchers proposed encoding answers with context, utilizing answer positions, or using text summaries for answer-aware or answer-agnostic QG \cite{sun2018answer,yuan2017machine,ma2020improving}.\\
In educational contexts, certain considerations must be taken when applying QG. Controlling question difficulty is crucial for effective education, with methods proposed to assess difficulty based on answerability, inference steps needed, or learners' abilities \cite{lord2012applications,uto2023difficulty}. Aligning questions with the syllabus is important for test focus, leading to studies training classifiers or ranking models to determine question relevance \cite{steuer2021not,hadifar2023diverse}.
Personalized education demands generating customized questions for students, prompting the development of knowledge-tracking models based on student answer histories or few-shot knowledge-tracking models incorporating sequences of student states and questions \cite{wang2023multi,srivastava2021question}.\\
However, none of the previous work in Question Generation was presented in Turkish. Here, we are presenting a new dataset \textit{Turkish-Quiz-Instruct} in Turkish and introducing different fine-tuned models for QG from the educational text in Turkish, including \textit{Llama-2-7b-chat-hf}, \textit{Llama-2-12b-chat-hf}, and \textit{GPT-3.5-Turbo}.

\section{Methodology}~\label{sec:method}
\begin{figure}[ht!]
    \centering
       \includegraphics[width=\textwidth]{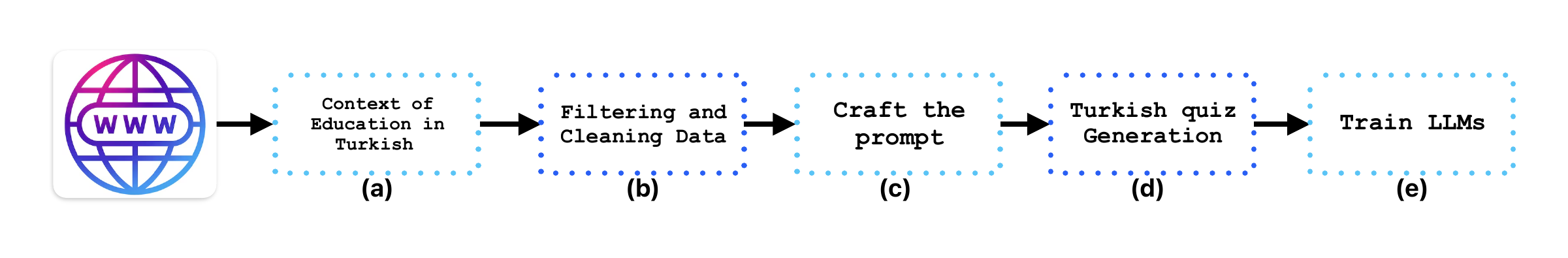}
    \caption{The diagram presents the methodology used in this study as follows:
(a) Data collection involving scraping educational content in Turkish, covering various subjects such as biology, history, etc.
(b) Data refinement and filtering to improve quality by removing overly short or excessively detailed content.
(c) Create prompts to generate Turkish quizzes based on the educational content.
(d) Utilization of \textit{GPT-4-Turbo} to produce quizzes from the collected data and configured prompts.
(e) Fine-tuning Large Language Models (LLMs) to generate Turkish educational quizzes from the given educational context.}
    \label{fig:fig1}
\end{figure}

In this study, we introduce the advancement of a Turkish educational quiz generator, leveraging state-of-the-art LLMs. We have curated a comprehensive dataset comprising Turkish educational texts \textit{Turkish-Quiz-Instruct} across various disciplines, including Chemistry, Biology, Geography,
Philosophy, Turkish Literature, and History. Subsequently, we have crafted multiple-choice and short-answer questions pertaining to these texts. In an iterative evaluation process, we refined our dataset to encompass a wide array of Turkish educational content, along with corresponding questions for each category.\\
To generate the Turkish quizzes and evaluate \textit{Turkish-Quiz-Instruct} dataset, we fine-tuned different LLMs under various scenarios, including both multiple-choice and short-answer formats. Among the models optimized were \textit{Llama-2-7b-chat-hf}, \textit{Llama-2-13b-chat-hf}, and \textit{GPT-3.5-Turbo}. This section will detail the methodologies employed in dataset generation and model tuning, illustrating the steps taken to achieve a functional Turkish quiz generator. Figure \ref{fig:fig1} illustrates the comprehensive methodology applied in this study.

\subsection{\textit{Turkish-Quiz-Instruct}}
In the preceding sections, we introduced a unique dataset that represents a significant advancement in the realm of Turkish educational resources. This carefully curated collection encompasses Turkish educational texts spanning various disciplines, such as Chemistry, Biology, Geography Philosophy, Turkish Literature, and History. A distinguishing feature of this dataset lies in its inclusion of corresponding questions presented in two distinct formats: multiple-choice-answer and short-answer questions. The creation of this dataset was a comprehensive process, executed with precision and illustrated in Figure \ref{fig:fig1}. Key stages involved in the dataset's development include scraping educational content from an array of online sources, systematically cleaning and filtering the obtained data to ensure quality and relevance, and thoughtfully designing prompts that would facilitate the generative process.\\
The generation of questions and answers was performed employing the advanced capabilities of \textit{GPT-4-Turbo}, a state-of-the-art LLM known for its exceptional performance in natural language understanding and production. This step was critical in enhancing the dataset's utility and applicability in educational settings by providing realistic and challenging questions aligned with the content. To ensure the high standard and educational value of the questions generated, an exhaustive evaluation process was conducted. This process scrutinized the questions for their accuracy, relevance, and pedagogical worth, ensuring they met the educational objectives.\\
In the subsequent sections, we will delve deeper into each of these stages, providing a comprehensive explanation that elucidates our methodology. This analysis will encompass the strategies employed for effectively scraping and filtering educational content, the rationale behind the prompt design, the technical details of leveraging \textit{GPT-4-Turbo} for generating educational material, and the criteria and measures taken during the evaluation phase. By unpacking these processes, we aim to offer insights into the meticulous approach undertaken to assemble a dataset that not only enriches Turkish educational resources but also sets a precedent for the development of similar datasets in other languages and disciplines.

\paragraph{Data Scraping}

The information extraction process commences with an initial filtering stage that targets a specific Turkish website dedicated to disseminating educational content for secondary school students. This repository encompasses a wide array of educational materials spanning various subjects, including mathematics, history, biology, and literature, among others. This study utilizes a dataset compiled from two primary online sources.\footnote{\url{https://www.https://bikifi.com/}\\ \url{https://www.https://basarisiralamalari.com/}}These resources offer comprehensive summaries of key concepts and topics aligned with officially sanctioned educational standards. The use of government-provided materials ensures alignment with established curricula and offers a standardized baseline. 

\paragraph{Data Cleaning and Filtering}
To ensure the integrity and quality of our dataset for subsequent analysis, a series of meticulous data-cleaning procedures were implemented. Specifically, extraneous elements such as emojis, links and excessively short text fragments were systematically eliminated. This meticulous cleansing process served to mitigate noise and irrelevant information, thereby enhancing the overall coherence and reliability of the data for further exploration.

\begin{figure*}[ht]
    \centering 
    \includegraphics[width=\textwidth]{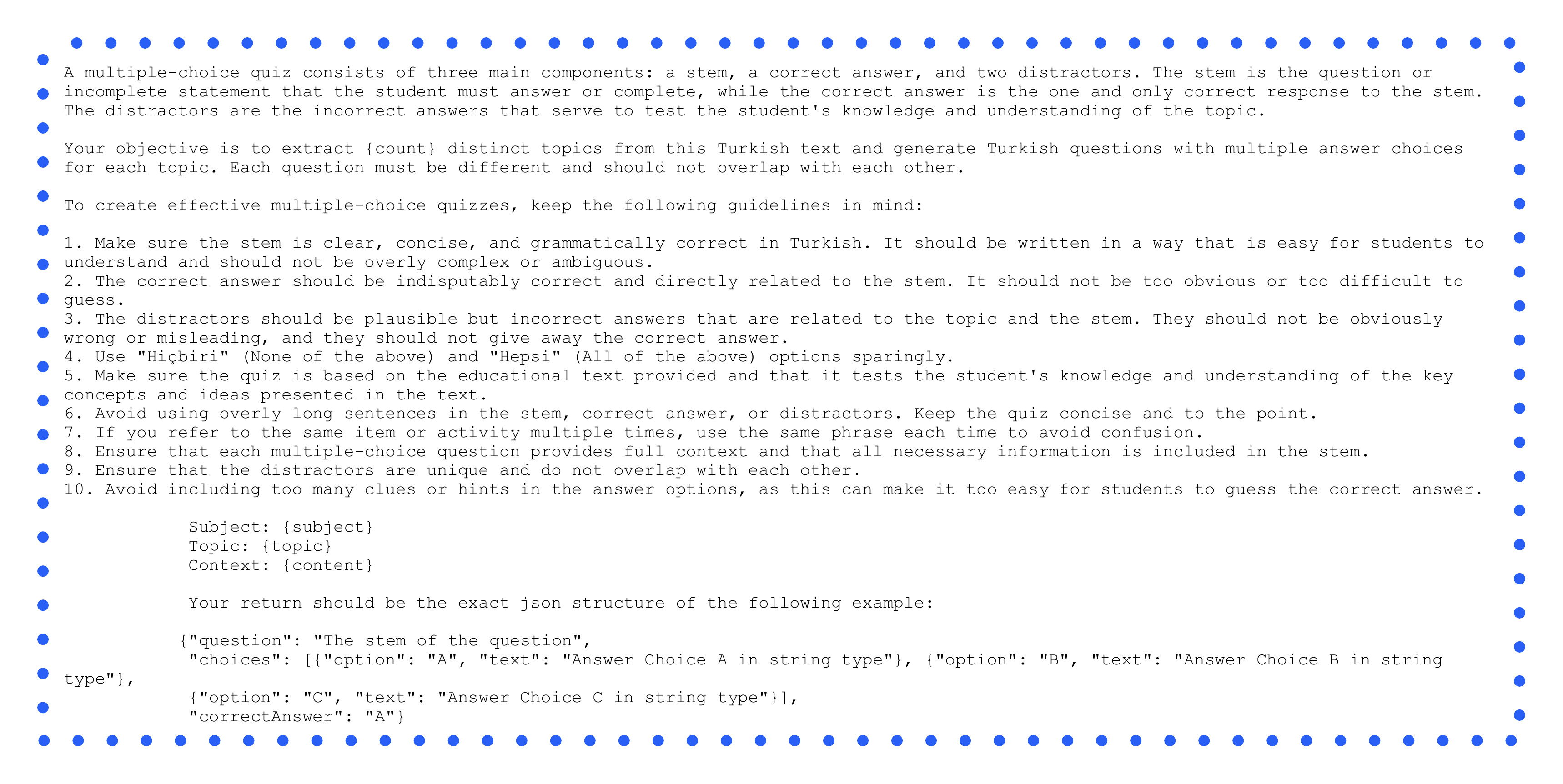}
    \caption{Turkish Quiz Generation Prompt.}
    \label{fig:fig2}
\end{figure*}

\paragraph{Craft the prompt.}
The formulation of targeted prompts was an integral part of our methodology, aimed at eliciting multiple-answer questions in Turkish from educational content. These prompts were meticulously crafted to guide the generation of questions that were both informative and engaging. Drawing on the topics and contexts provided by educational website pages, the prompts acted as navigational tools in the question-generation process. Our strategy was to create prompts uniquely tailored to each subject or topic, reflecting the specific nuances of the available information. The development of these prompts was essential to the success of our methodology, enhancing our capability to produce bespoke, high-quality Turkish quizzes for educational purposes. Figure \ref{fig:fig2} presents the prompt employed in this study.

\begin{figure*}[ht]
    \centering 
    \includegraphics[width=\textwidth]{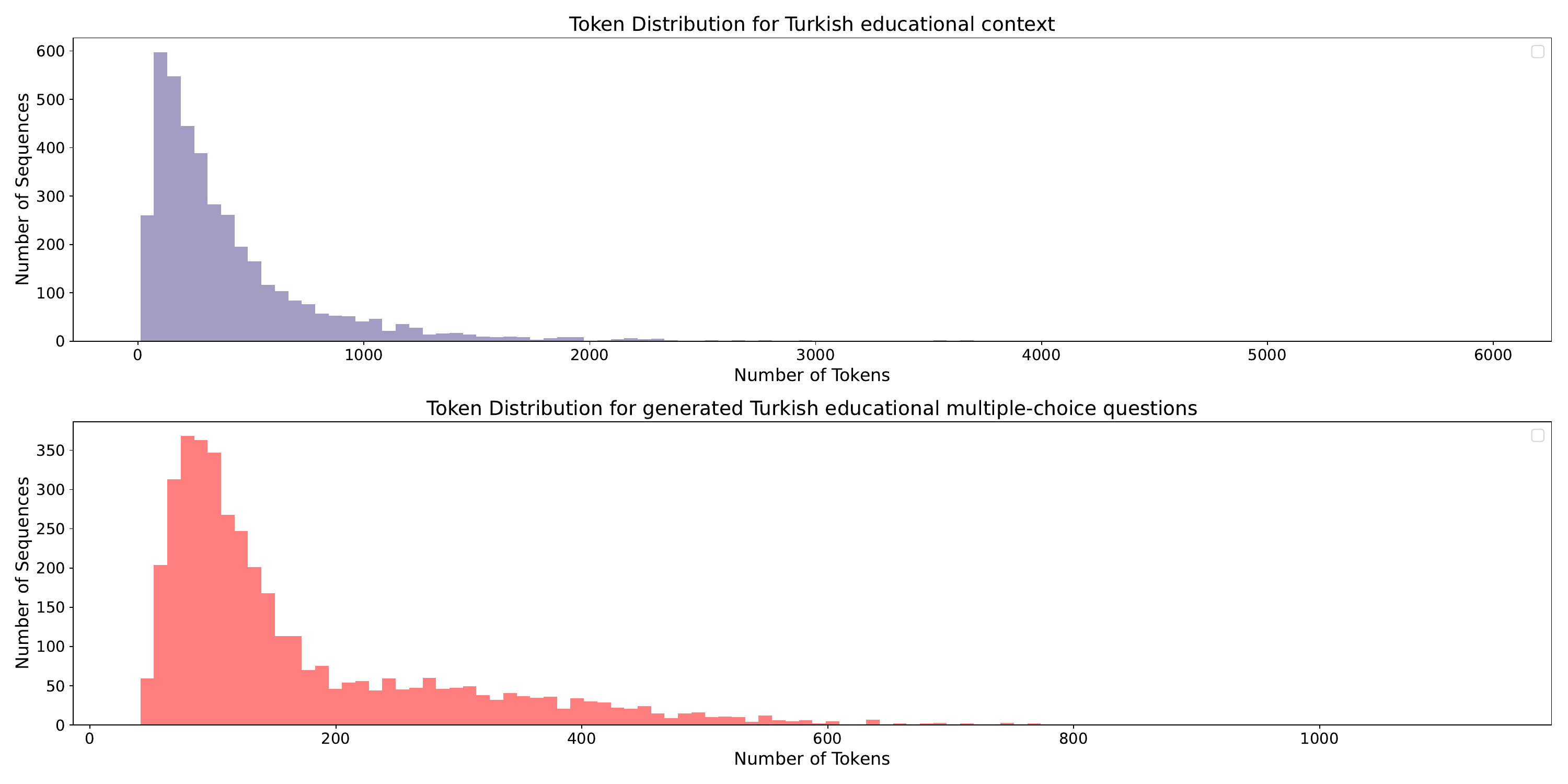}
    \caption{Token Distribution of Turkish educational context and generated Turkish educational multiple-choice questions using \textit{GPT-4-Turbo}}
    \label{fig:fig3}
\end{figure*}

\paragraph{Generating Educational multiple-answer questions.} \label{sec:datagen}
Our methodology, which capitalizes on the capabilities of Large Language Models (LLMs) for the autonomous generation of Turkish educational questions, is rooted in the principles outlined by the \textsc{self-instruct} framework, as explored \cite{wang2022self} central to our approach is the seamless integration of generated multiple-answer questions with contextual inputs, ensuring relevance and coherence. In this endeavor, \textit{GPT-4-Turbo} \cite{achiam2023gpt}, an advanced iteration renowned for its efficiency and performance, serves as our chosen LLM. This selection reflects our strategic process of leveraging meticulously curated content, topics, and prompts, thereby fabricating a customized foundation conducive to producing finely tuned multiple-answer questions tailored to educational objectives. \ref{fig:fig3} displays the distribution of tokens within the generated Turkish educational multiple-answer questions, in this plot we used the fast \textit{Llama-2} tokenizer.

\begin{figure}[ht!]
    \centering
    \includegraphics[width=1\textwidth]{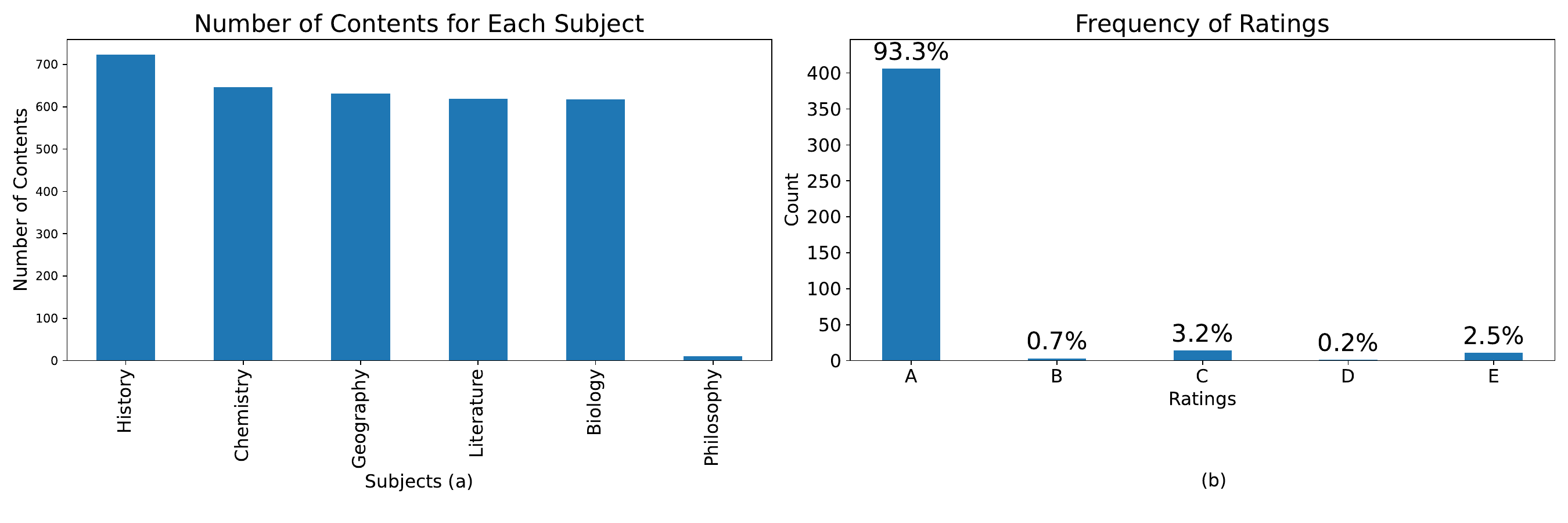}
    \caption{(a) Content Subject Distribution and (b) GPT-4 Ratings from the Human Evaluation}
    \label{fig:combined}
\end{figure}

In Figure \ref{fig:combined} (a), the distribution of various educational subjects within the \textit{Turkish-Quiz-Instruct} dataset is illustrated.

\paragraph{Evaluating Generated Data Quality}

The evaluation of generated Turkish educational quizzes faces significant obstacles due to the lack of a reference corpus, which is essential for benchmarking these quizzes using metrics like ROGUE scores. This absence challenges the process of assessing the quality of educational quiz generation. However, the unique nature of the task, where the creation of effective questions necessitates the subtle rewording of reference texts, necessitates an evaluation method focused on high levels of textual extraction.\\
In response to this, our methodology incorporates the use of ROUGE-L scores to compare the content of generated questions with the input text, aiming to quantify the degree to which these questions remain faithful to the original context. A high ROUGE-L score suggests a significant fidelity to the context, serving as a measure to both deter the generation of irrelevant content and discourage the creation of overly simplistic quizzes or the inadvertent inclusion of answers within the questions. The results from this methodology have been promising, with an achieved average ROUGE-L score of 22 indicating a correlation between the questions and the context's corresponding sentences.

    

Complementing the quantitative assessment, a qualitative examination was also undertaken through human evaluation. The evaluation of the generated quizzes employed a human-centric approach. Native Turkish speakers with demonstrable expertise in the linguistic subtleties of the language were recruited to act as judges. This ensured a rigorous assessment process grounded in an understanding of the nuances that influence effective quiz construction and evaluation within the Turkish language. A representative sample of questions underwent scrutiny by a panel of Turkish language specialists, adopting a rigorous evaluation framework akin to the one described in \cite{wang2022self}, which utilizes a five-point rating scale. This comprehensive approach, melding both quantitative and qualitative evaluations, significantly enhances the analysis of the efficiency and effectiveness of the generated Turkish educational quizzes. following is the five-point rating scale which we used in this study:

\begin{itemize}
    \item \texttt{RATING-A}: Questions are factually accurate and directly relate to significant concepts from the source text.
    \item \texttt{RATING-B}: Questions are mostly factual and related to the text, however, there may be lapses in directly addressing significant concepts or some minor grammatical issues.
    \item \texttt{RATING-C}: Questions are loosely related to the text but may address topics.
    \item \texttt{RATING-D}: Questions contain factual inaccuracies or are only minimally relevant to the text.
    \item \texttt{RATING-E}: Questions are demonstrably wrong or misleading or have no clear connection to the educational text.
\end{itemize}
The results of the human evaluation are presented in Figure \ref{fig:combined} (b), where it is evident that a significant number (93.3\%) of the generated questions were rated as \texttt{RATING-A}. This rating indicates the high quality of the newly introduced dataset.

\subsection{From LLMs to Turkish Educational Quizzes}
To generate Turkish educational questions from educational Turkish textual sources and evaluate the proposed dataset \textit{Turkish-Quiz-Instruct}, we fine-tuned LLMs including \textit{Llama-2-7b-chat-hf}, \textit{Llama-2-13b-chat-hf}, and \textit{GPT-3.5-Turbo} \cite{brown2020language,touvron2023llama}, chosen for their support of the Turkish language.\\
Before applying these models to question generation, they underwent an extensive fine-tuning process. This step was crucial and benefited from a carefully curated dataset \textit{Turkish-Quiz-Instruct} which we introduced in the previous section, rich in educational content pertinent to our objectives, as detailed in the preceding section. This dataset acted as a foundation for adapting the models, enhancing their ability to produce questions that are contextually relevant and linguistically precise in Turkish.\\
The fine-tuning methodology was thorough, entailing iterative tuning of the model using Parameter Efficient Fine-Tuning to reduce task-specific loss. This procedure aimed not just at refining the models' comprehension of educational material but also at ensuring the nuanced representation of the Turkish language in the output questions. Achieving high fidelity in language generation, given the content diversity and linguistic complexity, was a challenging endeavor.\\
Through meticulous customization of state-of-the-art LLMs using the focused dataset \textit{Turkish-Quiz-Instruct}, we markedly advanced their Turkish question-generating capabilities. This strategy guaranteed that the generated questions were pertinent and stimulating and conformed to Turkish educational criteria linguistically and pedagogically.

\section{Experiments}~\label{sec:experiments}
This section describes the methodology employed for fine-tuning LLMs to enhance their performance on the Turkish quiz generation task. The \textit{Turkish-Quiz-Instruct} dataset, constructed as detailed in Section \ref{sec:method}, served as the foundation for this process. Our approach involved manipulating this dataset in two distinct ways to facilitate the generation of varied question formats, specifically multiple-choice questions (MCQs) and short-answer questions (SAQs).\\
Initially, the dataset was designed to encompass multiple-choice questions, to adapt this format for generating short-answer questions, we extracted the correct answers from the MCQs while discarding the incorrect options. This manipulation resulted in two separate datasets tailored to each question type.\\
Subsequently, we fine-tuned three different LLMs - \textit{GPT-3.5-Turbo}, \textit{Llama-2-7b-chat-hf}, and \textit{Llama-2-13b-chat-hf} - utilizing these datasets. The objective was to assess the models' proficiency in generating two types of questions from Turkish educational texts: multiple-choice and short-answer questions. The following sections provide a detailed examination of the experiments conducted for each question format, leveraging the distinct datasets and LLMs mentioned.\\
For Fine-Tuning, The \textit{Turkish-Quiz-Instruct} dataset was then partitioned into two sets: one comprising 8000 texts designated for training purposes, and the other containing 260 texts allocated for evaluation using automatic metrics and human evaluation.

\subsection{Fine-Tuning Configuration}

We undertook the fine-tuning of three advanced models: \textit{GPT-3.5-Turbo}, \textit{Llama-2-7b-chat-hf}, and \textit{Llama-2-13b-chat-hf}. The fine-tuning process for \textit{GPT-3.5-Turbo} was executed with a batch size of $16$ and a learning rate of $0.001$ over three epochs. In contrast, the fine-tuning of both \textit{Llama-2} models leveraged Parameter Efficient Fine-Tuning (PEFT) techniques, with settings of $r=16$ and $\alpha=32$ and a unified batch size of $64$, applying a learning rate of $0.0001$ across three epochs for each model.

\subsection{Turkish educational multiple-choice questions generation}
In the pursuit of extracting relevant insights within the scope of specific multiple-choice questions and subjects from a Turkish Educational text corpus, our methodology began with the articulation and subsequent refinement of the \textit{Turkish-Quiz-Instruct} dataset, as detailed in Section \ref{sec:method}.\\
The subsequent phase of our research entailed a rigorous evaluation of the models' capabilities on the reserved set of evaluation texts. Initially, this evaluation leveraged prominent metrics such as Rouge 1, Rouge 2, and Rouge L, facilitating a comparison between the question quality emanating from our fine-tuned models and that of the \textit{Turkish-Quiz-Instruct} You can review the results in Table \ref{tab:llms_results-multi}, which demonstrate that the Rouge scores across all models have improved following fine-tuning. Considering that ROUGE scores are not effective metrics for evaluating the generated Turkish questions, it is important to note some limitations. Firstly, certain questions might be relevant despite deviating from the content found in the \textit{Turkish-Quiz-Instruct} dataset. Additionally, ROUGE scores do not account for syntactic and grammatical errors that may be present in the questions, further undermining their reliability as evaluative tools, so This comparative analysis was enriched by the interpretations of human evaluators. Furthermore, an additional layer of analysis was applied to juxtapose the performances of the models in their base (pre-fine-tuning) form against their post-fine-tuning renditions, employing the same established five-level rating for a thorough assessment in the section \ref{sec:datagen}. The extensive outcomes from these evaluative stages are scrupulously documented, and accessible in Table \ref{fig:llms-ratings-multi1} for an enriched understanding of the findings. Reviewing the data presented in Figure \ref{fig:llms-ratings-multi1}, it is evident that the performance of all models has improved post-fine-tuning. Notably, \textit{GPT3.5-Turbo} displayed a significant increase in the number of \texttt{RATING-A} assessments. Moreover, initially, both \textit{Llama-2-7b-chat-hf} and \textit{Llama-2-13b-chat-hf} struggled with generating Turkish multiple-answer questions from provided texts, with all responses categorized as \texttt{RATING-E}. However, after fine-tuning using the \textit{Turkish-Quiz-Instruct} dataset, these models demonstrated enhanced capabilities. Specifically, \textit{Llama-2-13b-chat-hf} showed a remarkable rise in receiving \texttt{RATING-A}, indicating substantial improvements.
 \begin{figure}[ht!]
    \centering
    \includegraphics[width=\textwidth]{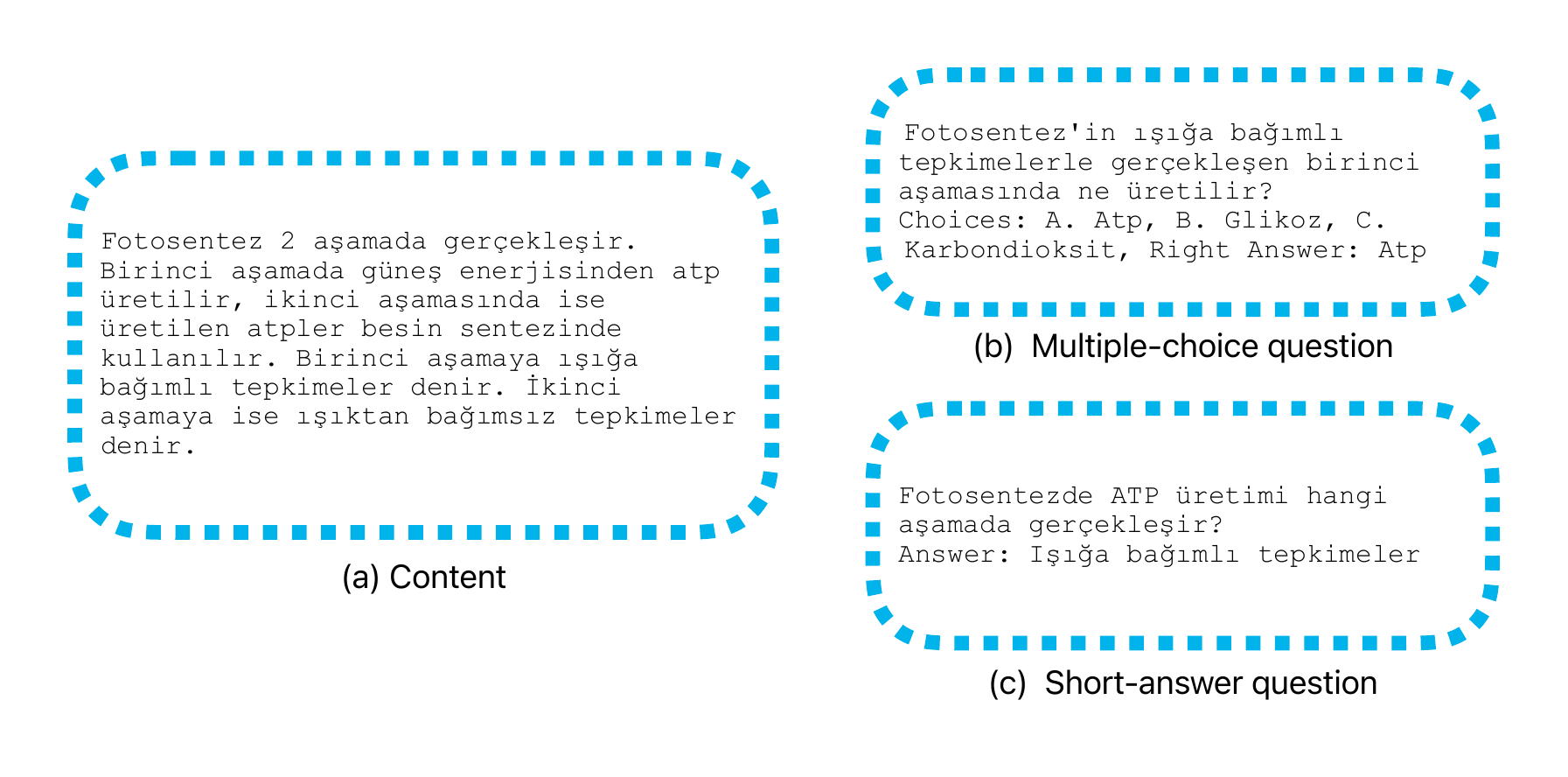}
    
    \caption{Sample Questions Created by Fine-Tuned Language Models.}
    \label{fig:example-questions}
\end{figure}

 \begin{figure}[ht!]
    \centering
    \includegraphics[width=\textwidth]{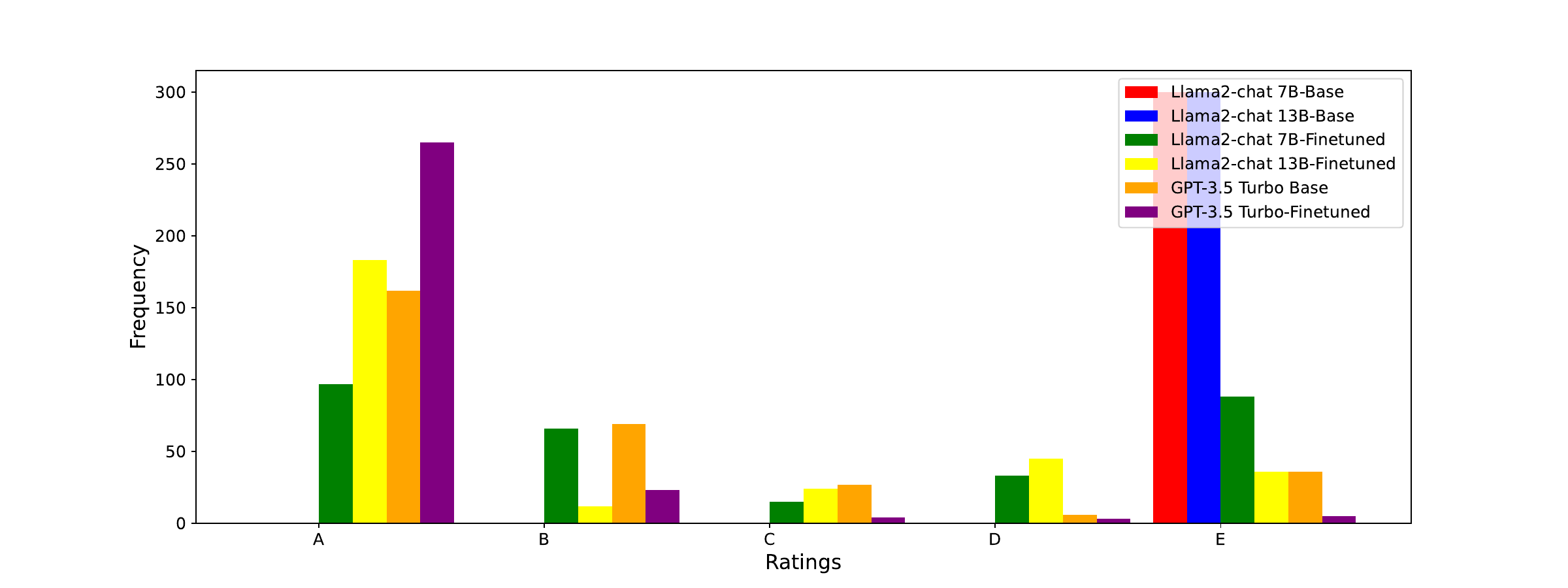}
    
    \caption{Human evaluation of the performance of LLMs with and without fine-tuning for MCQs.}
    \label{fig:llms-ratings-multi1}
\end{figure}
 \begin{figure}[ht!]
    \centering
    \includegraphics[width=\textwidth]{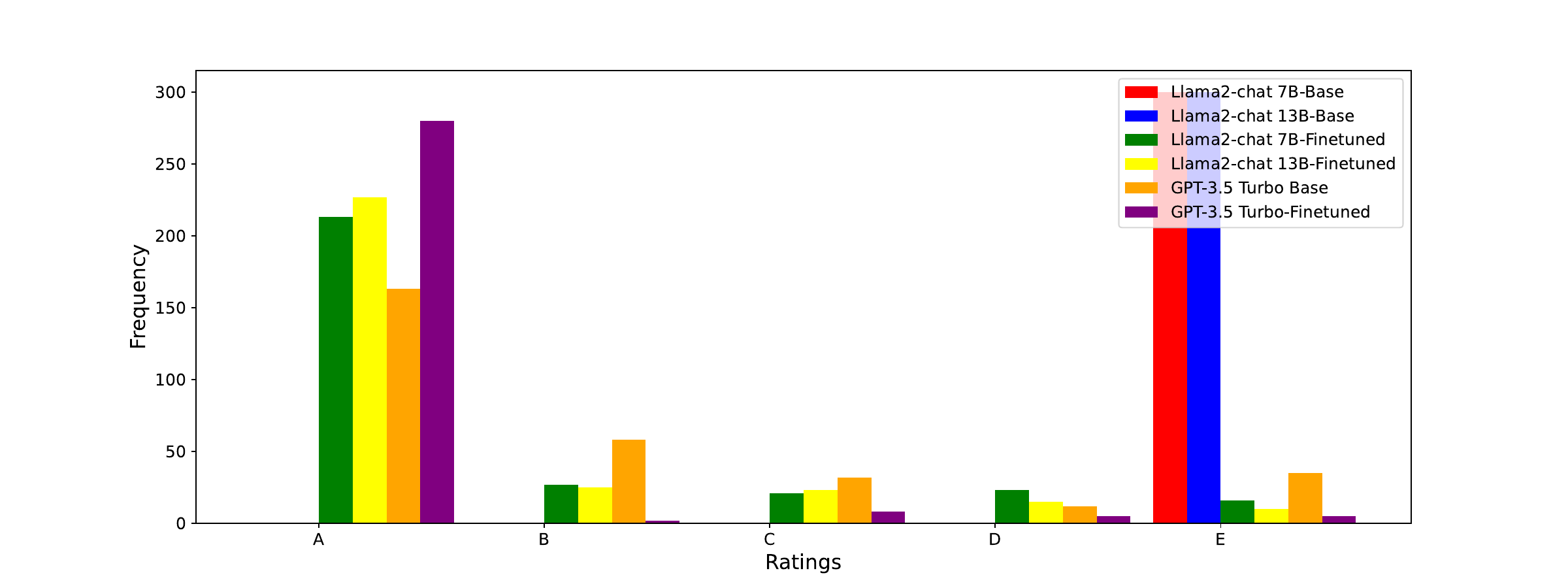}
    
    \caption{Human evaluation of the performance of LLMs with and without fine-tuning for SAQs.}
    \label{fig:llms-ratings-short1}
\end{figure}

\begin{table}[ht!]
     \centering
     \small
     \begin{tabular}{cccccc}
      \hline
    \textbf{model type}&    \textbf{model name} & \textbf{\# params} & \textbf{ROUGE-1}& \textbf{ROUGE-2} & \textbf{ROUGE-L} \\ \hline
     &\textsc{Llama2-chat}  & 7B & 4.71 & 1.42 & 3.88\\
   Base LLMs &\textsc{Llama2-chat}  & 13B & 5.33 & 1.73 & 4.28\\
    &\textsc{GPT3.5 Turbo}  & - & 27.53 & 13.18 & 21.38\\
     \hline
   
    &\textsc{Llama2-chat}  & 7B & 17.33 & 2.88 & 12.11\\
   Finetuned LLMs
    &\textsc{Llama2-chat}  & 13B & 24.34 & 11.44 & 19.75\\
    
    &\textsc{GPT3.5 Turbo}  & - & \textbf{36.64} & \textbf{20.36} & \textbf{28.82}\\
    \hline
     \end{tabular}
     \caption{Comparing rouge scores for pre- and post-fine-tuned LLMs with \textit{Turkish-Quiz-Instruct} dataset for MCQs.}
     \label{tab:llms_results-multi}
 \end{table}
\subsection{Turkish short-answer questions generation}
Here, we explored the utility of short-answer question generation from given Turkish texts, with an emphasis on fine-tuning to enhance the precision of the generated questions, we fine-tuned the discussed models for this task using the \textit{Turkish-Quiz-Instruct} dataset.\\
Our evaluation framework focused on the comparative analysis of the generated questions' quality, utilizing both, base models (before fine-tuning) and the fine-tuned versions. We employed the standard metrics for this purpose, namely Rouge 1, Rouge 2, and Rouge L. These metrics facilitated a structured comparison against the performance of \textit{GPT-4-Turbo}'s generated questions which construct the \textit{Turkish-Quiz-Instruct} dataset, Due to the limitations of the Rouge Score as we discussed earlier, human evaluators also scrutinized the model's performance. The outcomes of these comparative assessments, highlighting the improvement in question quality attributable to our fine-tuning efforts, are meticulously documented in Table \ref{tab:llms_results-short} and Figure \ref{fig:llms-ratings-short1} for detailed examination. in Table \ref{tab:llms_results-short} you can see the Rouge scores improved after fine-tuning for all of the models.
\begin{table}[ht!]
     \centering
     \small
     \begin{tabular}{cccccc}
      \hline
    \textbf{model type}&    \textbf{model name} & \textbf{\# params} & \textbf{ROUGE-1}& \textbf{ROUGE-2} & \textbf{ROUGE-L} \\ \hline
     &\textsc{Llama2-chat}  & 7B & 3.31 & 1.04 & 2.86\\
   Base LLMs &\textsc{Llama2-chat}  & 13B & 3.81 & 1.33 & 3.22\\
    &\textsc{GPT3.5 Turbo}  & - & 28.19 & 15.87 & 22.22\\
     \hline
   
    &\textsc{Llama2-chat}  & 7B & \textbf{54.37} & \textbf{47.33} & 49.01\\
   Finetuned LLMs
    &\textsc{Llama2-chat}  & 13B & 52.99 & 46.88 & \textbf{49.54}\\
    
    &\textsc{GPT3.5 Turbo}  & - & 45.31 & 31.59 & 40.32\\
    \hline
     \end{tabular}
     \caption{Comparing rouge scores for pre- and post-fine-tuned LLMs with \textit{Turkish-Quiz-Instruct} dataset for SAQs.}
     \label{tab:llms_results-short}
 \end{table}

Given the unreliability of Rouge scores for accurately assessing the true performance of our models, we incorporated human evaluations into our assessment framework. We utilized the same five-level rating and conditions from our annotation of the generated data by \textit{GPT4-Turbo}, discussed in Section \ref{sec:datagen}. The results are presented in Figure \ref{fig:llms-ratings-short1}, where it becomes evident that both \textit{Llama-2-7b-chat-hf} and \textit{Llama-2-13b-chat-hf} initially struggled to generate any acceptable Turkish short-answer questions before fine-tuning. However, post-fine-tuning these models demonstrated a substantial improvement. In Figure \ref{fig:llms-ratings-short1}, before fine-tuning, all questions generated by  \textit{Llama-2-7b-chat-hf} and  \textit{Llama-2-13b-chat-hf}, were rated as \texttt{RATING-E}, and these models were incapable of generating questions in Turkish. After fine-tuning, however, there is a significant improvement in the number of questions rated \texttt{RATING-A}. Notably, the \textit{Llama-2-13b-chat-hf} model displays superior performance over the \textit{Llama-2-7b-chat-hf}, despite the latter having fewer parameters. Similarly, \textit{GPT3.5-Turbo} exhibited improved performance in generating Turkish short-answer questions once fine-tuned with the introduced dataset \textit{Turkish-Quiz-Instruct}, as evidenced by the significant rise in \texttt{RATING-A}, underscoring the quality of the dataset.

\section{Conclusion}~\label{sec:conclusions}
In this comprehensive study, we have successfully introduced the Turkish educational quiz generator, marking a significant advancement in the field of educational technology in the Turkish language. To cater to diverse learning preferences and testing requirements, we have developed and incorporated two distinct types of questions: multiple-answer and short-answer questions. Additionally, we have established the inaugural dataset named \textit{Turkish-Quiz-Instruct} for educational quizzes in Turkish, meticulously curated to support these question formats, thereby addressing a notable gap in available resources for Turkish-language education.\\
To enhance the practical utility and effectiveness of our quiz generator, we meticulously formatted the dataset in two variations: multiple-answer and short-answer, enabling educators and technology developers to leverage this tool for a broad spectrum of educational applications. Taking advantage of this unique dataset, we have undertaken the fine-tuning of several leading Large Language Models (LLMs), including\textit{Llama-2-7b-chat-hf}, \textit{Llama-2-13b-chat-hf}, and \textit{GPT-3.5-Turbo}. Our objective was to adapt these models for the generation of Turkish educational questions directly from educational text materials, thereby simplifying and enriching the process of quiz content creation.\\
Our findings reveal a marked improvement in the models' performance post-fine-tuning. Initially, these sophisticated LLMs struggled to generate coherent and contextually appropriate Turkish questions from educational texts, demonstrating a significant language and domain-specific knowledge gap. However, through the strategic application of our specially designed dataset for fine-tuning, we observed a dramatic enhancement in their capability to create meaningful and accurate quiz questions in Turkish.\\
The implications of this study are far-reaching, offering a promising avenue for educators, technologists, and linguists to develop interactive, engaging, and linguistically diverse educational tools. This not only democratizes access to high-quality education in the Turkish language but also sets a precedent for similar innovations in other languages and domains. Our research underscores the critical role of specialized datasets and the potential of LLM fine-tuning in bridging linguistic and educational divides, paving the way for a more inclusive and effective global education ecosystem. For future work, we aim to enhance language models for the Turkish context and then specialize them to generate various types of questions.

\end{document}